\documentclass[sn-mathphys-num]{sn-jnl}

\usepackage{graphicx}
\usepackage{multirow}
\usepackage{amsmath,amssymb,amsfonts}
\usepackage{amsthm}
\usepackage{mathrsfs}
\usepackage[title]{appendix}
\usepackage{xcolor}
\usepackage{textcomp}%
\usepackage{manyfoot}%
\usepackage{booktabs}%
\usepackage{algorithm}%
\usepackage{algorithmicx}%
\usepackage{algpseudocode}%
\usepackage{listings}%
\usepackage{bm}
\usepackage{tabularray}
\usepackage{pifont}
\usepackage{setspace}
\usepackage{makecell}
\usepackage{colortbl}
\usepackage{hyperref}
\definecolor{tabcolor}{rgb}{.105,.410,.113}
\definecolor{mygray}{gray}{.5}

\theoremstyle{thmstyleone}%
\theoremstyle{thmstyletwo}%

\theoremstyle{thmstylethree}%

\begin{document}

\title[Article Title]{AdaBrain-Bench: Benchmarking Brain Foundation Models for Brain-Computer Interface Applications}



\author[1,2]{\fnm{Jiamin} \sur{Wu}}
\equalcont{These authors contributed equally to this work.}

\author[1]{\fnm{Zichen} \sur{Ren}}
\equalcont{These authors contributed equally to this work.}

\author[1,3]{\fnm{Junyu} \sur{Wang}}

\author[1,4]{\fnm{Pengyu} \sur{Zhu}}

\author[1,3]{\fnm{Yonghao} \sur{Song}}

\author[1]{\fnm{Mianxin} \sur{Liu}}

\author[1]{\fnm{Qihao} \sur{Zheng}}

\author[1]{\fnm{Lei} \sur{Bai}}

\author*[1,2]{\fnm{Wanli} \sur{Ouyang}}

\author*[1]{\fnm{Chunfeng} \sur{Song}} \email{songchunfeng@pjlab.org.cn}


\affil[1]{\orgname{Shanghai Artificial Intelligence Laboratory}}

\affil[2]{\orgname{The Chinese University of Hong Kong}}

\affil[3]{\orgname{Tsinghua University}}

\affil[4]{\orgname{North China Electric Power University}}




\abstract{Non-invasive Brain-Computer Interfaces (BCI) offer a safe and accessible means of connecting the human brain to external devices, with broad applications in home and clinical settings to enhance human capabilities. However, the high noise level and limited task-specific data in non-invasive signals constrain decoding capabilities. Recently, the adoption of self-supervised pre-training is transforming the landscape of non-invasive BCI research, enabling the development of brain foundation models to capture robust neural representations from large-scale unlabeled electroencephalography (EEG) signals. However, despite these advances, the field currently lacks standard and comprehensive benchmarks to assess the utility of the public foundation models across diverse BCI tasks, hindering their widespread adoption. To address this challenge, we present AdaBrain-Bench, a large-scale standardized benchmark to systematically evaluate brain foundation models in widespread non-invasive BCI tasks. AdaBrain-Bench encompasses a diverse collection of representative BCI decoding datasets spanning 7 key applications. It introduces a streamlined model adaptation pipeline integrated with multi-dimensional evaluation metrics and a set of adaptation tools for flexible deployment. The benchmark establishes multiple transfer settings, including cross-subject, multi-subject, and few-shot settings, to rigorously evaluate models' transferability across diverse real-world application scenarios. We leverage AdaBrain-Bench to evaluate publicly available brain foundation models and offer insights into practices for selecting appropriate models in various scenarios. We make our benchmark pipeline available at \href{https://github.com/Jamine-W/AdaBrain-Bench}{GitHub repository} to enable reproducible research and external use, offering a continuously evolving platform to foster progress toward robust and generalized neural decoding solutions.
}

\maketitle

\section{Introduction}

Brain-Computer Interfaces (BCIs) have sparked significant interest in creating systems that facilitate direct information transmission between human brains and computers or external devices~\cite{Lemm, Cecotti, Deneen}. 
Over recent decades, several teams have used BCIs to efficiently decode movement~\cite{flesher2021brain, ding2025eeg} and communication~\cite{anumanchipalli2019speech, willett2023high, tang2023semantic, chen2024neural, wairagkar2025instantaneous} from electrodes implanted in the cortex or over its surface.
Despite their efficacy, the requirement for invasive brain surgery limits the applicability of these decoding methods and these high-quality signals are difficult to maintain chronically.
Several laboratories have thus focused on non-invasive BCIs that decode brain activity from non-invasive recordings. The most popular non-invasive BCI technique is
electroencephalography (EEG). EEG provides millisecond-level monitoring of macroscopic changes of electric signals elicited in the cortex~\cite{Ramadan, Ang, Noise-factorized}, offering a safe and potentially wearable approach for neural data acquisition.
Brain decoding techniques~\cite{Channel-Projection} that interpret latent neural patterns underlying EEG signals have powered a diverse range of BCI applications, spanning motor imagery decoding~\cite{Ding2025}, sleep stage classification~\cite{DeepSleepNet}, seizure detection~\cite{Acharya}, mental state classification~\cite{graph, workload} and visual decoding~\cite{NEURIPS2024_ba5f1233}.

\begin{figure}[tbp]
\centering
\includegraphics[width=1.0\textwidth]{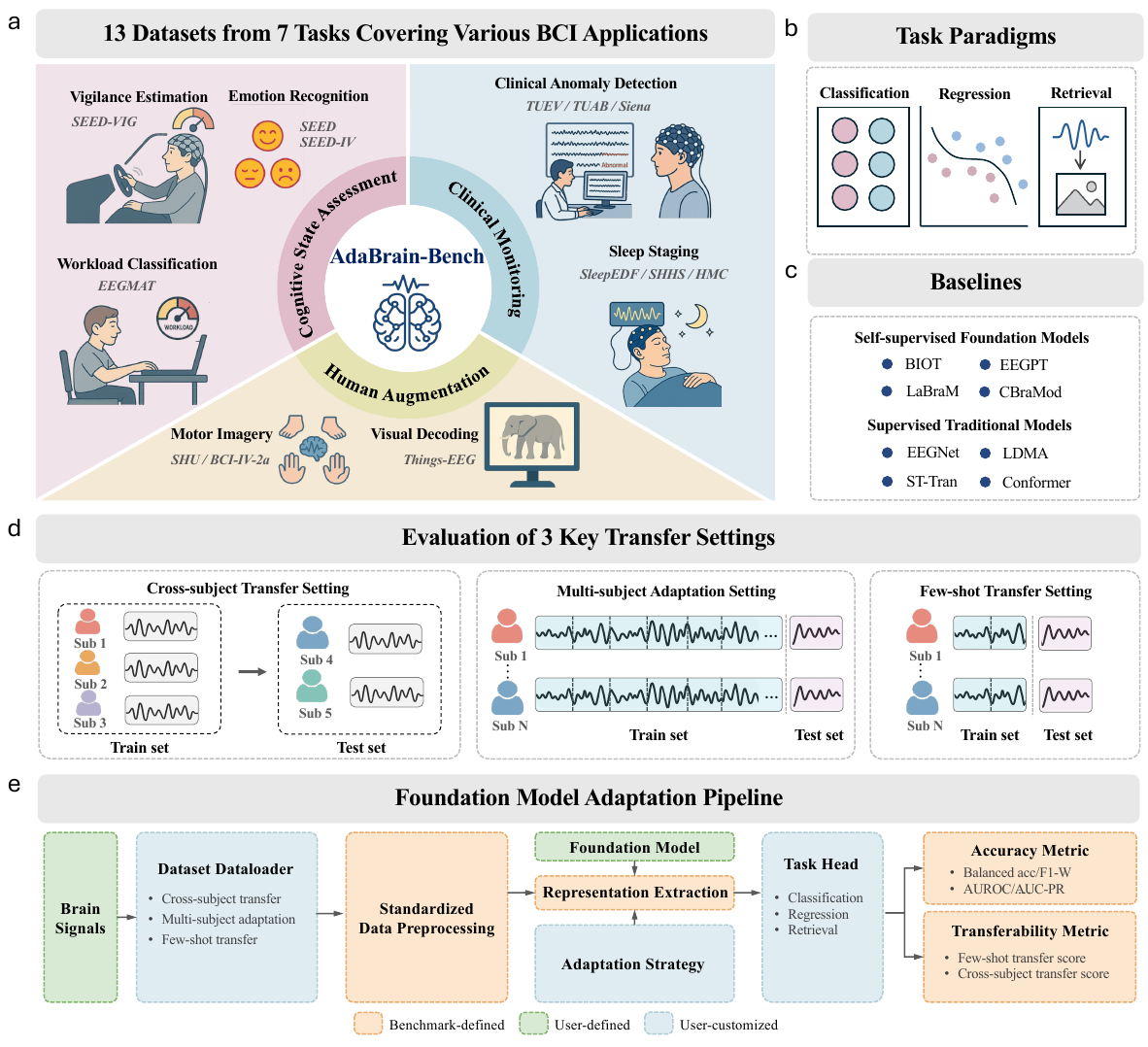}
\caption{
\textbf{Overview of AdaBrain-Bench}. \textbf{a}, categorization of 13 benchmark datasets and 7 benchmark tasks ranging from cognitive state assessment, human augmentation, and clinical monitoring. \textbf{b}, diverse task paradigms including classification, regression, and retrieval paired with comprehensive evaluation metrics. \textbf{c}, evaluated baselines including several recently open-sourced self-supervised foundation models (BIOT~\cite{yang2023biot}, EEGPT~\cite{eegpt}, LaBraM~\cite{labram}, and CBraMod~\cite{wang2024cbramod}) and supervised traditional models (EEGNet~\cite{Lawhern}, LDMA~\cite{miao2023lmda}, ST-Tran~\cite{Transformer}, and Conformer~\cite{Conformer}) used for comparison. \textbf{d},~multifaceted evaluation settings including cross-subject transfer, multi-subject adaptation and few-shot transfer settings to thoroughly assess models downstream task generalization ability in various scenarios. \textbf{e},~the overall foundation model adaptation pipeline. Users input their model to the benchmark and brain signal data from their selected task.
}
\label{overall}
\end{figure}

Traditional brain decoding efforts utilize supervised deep learning techniques for various BCI tasks, including convolution neural networks (CNN)~\cite{Cecotti, Sakhavi, Fatigue}, long short-term memory (LSTM)~\cite{LSTM, Tsiouris, Emotion}, and Transformer~\cite{Conformer, SleepTransformer, Transformer}.
Nevertheless, EEG recordings exhibit an inherent low signal-to-noise ratio and present significant variability~\cite{schirrmeister2017deep, defossez2023decoding}.
Moreover, compared to image or text data, the high cost of EEG data collection and annotation makes it challenging to build large, high-quality, task-specific  datasets, resulting in data scarcity for certain applications~\cite{labram, zhao2020deep}.
%
%
These challenges make traditional supervised deep learning-based decoding methods struggle in learning effective neural representation~\cite{zhang2021hybrid, zhao2020deep}, and limit their generalization ability for real-world BCI deployment. 

%

To address these challenges, recent efforts have been focusing on adopting self-supervised pre-training (SSP) to revolutionize brain decoding pipelines.
Self-supervised pre-training algorithms have driven a paradigm shift by enabling deep neural networks to be trained on vast unlabeled datasets, achieving performance on par with supervised learning. Large neural networks trained this way can be described as foundation models that can be used for diverse downstream tasks with minimal fine-tuning. 
SSP is well-suited for developing robust and generalizable brain decoding models, as it enables training brain foundation models on large unannotated EEG data from diverse sources.
Therefore, SSP for brain decoding has recently garnered attention, with increasing efforts~\cite{labram, yang2023biot, eegpt, wang2024cbramod} focused on developing brain foundation models for generalized BCI decoding. Table~\ref{tab:summary-model} provides a summary of open-source brain foundation models.
%
%
%
For instance, BIOT~\cite{yang2023biot} proposes a unified tokenizer to encode EEG signals of varying forms into structured segments, enabling efficient cross-data learning. 
LaBraM~\cite{labram} attempts to learn robust EEG representations through masked signal modeling of discretized EEG tokens during pretraining, achieving consistently enhanced decoding performance after pretrained on 2,500 hours of EEG data.  

\begin{table*}[tbp]
\centering
\caption{\textbf{A comparative analysis from representative studies assessing EEG-based brain decoding.} The compared elements involve evaluated models (including self-supervised foundation and supervised tradition models), the evaluation scope (including tasks from cognitive state assessment, human augmentation, and clinical monitoring), the evaluation settings (cross-subject, multi-subject and few-shot settings) and adaptation strategies (full fine-tune and linear probe).}
\label{tab:study-comparison}
\small
\renewcommand{\arraystretch}{1.5}
\setlength{\tabcolsep}{2.5pt}{\resizebox{0.99\textwidth}{!}{
\begin{tabular}{c|cc|c|cc|ccc|ccc|cc}
\toprule
\multirow{3}{*}{\textbf{Study}} 
&
\multicolumn{2}{c|}{\multirow{2}{*}{\textbf{Evaluated Models}}}
& \multicolumn{6}{c|}{\textbf{Evaluation Scope}} & \multicolumn{3}{c|}{\multirow{2}{*}{\textbf{Evaluation Settings}}} & \multicolumn{2}{c}{\multirow{2}{*}{\makecell{\textbf{Fine-tune}\\ {\textbf{Strategy}}}}} \\
\cmidrule[0.1pt]{4-9}
 & & & \makecell{\textbf{Cognitive State} \\\textbf{Assessment}} & \multicolumn{2}{c|}{\makecell{\textbf{Human} \\ \textbf{Augmentation}}} & \multicolumn{3}{c|}{\textbf{Clinical Monitoring}} & &&&&
\\ 
\cmidrule[0.1pt](lr){2-14} 
 & \makecell{Foundation\\Models}
 & \makecell{Traditional\\ Models}
&\makecell{Workload\\Classification} & \makecell{Motor\\Imagery} & \makecell{Visual\\Decoding} &
  \makecell{Emotion\\Recognition} &
 \makecell{Clinical Anomaly
\\Detection}  & \makecell{Sleep\\Staging} & \makecell{Cross-\\subject} & \makecell{Multi-\\subject} & \makecell{Few-\\shot} & \makecell{Full Fine-\\tune} & \makecell{Linear\\Probe}
\\
\midrule
\textbf{ALBM}~\cite{assessing} & \ding{51} & \ding{51} & &\ding{51}& & & & \ding{51} & \ding{51} & & & &
\\
\textbf{LibEER}~\cite{liu2024libeer} & & \ding{51} & & & & \ding{51}&  & & \ding{51}& & & &
\\
\textbf{SzCORE}~\cite{dan2024szcore} &  & \ding{51} & & & & & \ding{51} & & \ding{51} & & & &
\\
\textbf{LaBraM}~\cite{labram}  
 &  \ding{51} &  \ding{51} &  & \ding{51} &  & \ding{51} & \ding{51}  & & \ding{51} & \ding{51} &  & \ding{51} & \ding{51}
\\
\textbf{CbraMod}~\cite{wang2024cbramod} & \ding{51}& \ding{51} & \ding{51} & \ding{51} &  & \ding{51} & \ding{51}  & \ding{51} & \ding{51} & \ding{51} &	 &	\ding{51} & 
\\
\textbf{EEGPT}~\cite{eegpt} & \ding{51}& \ding{51} &  & \ding{51} &  &  & \ding{51} &  \ding{51} & \ding{51} & \ding{51} &  & \ding{51} & \ding{51}
\\
\textbf{BIOT}~\cite{yang2023biot} & & \ding{51}&  &  &  &  & \ding{51}  &  & \ding{51} &  &  & \ding{51} & 
\\
\textbf{MOABB}~\cite{chevallier2024largest} & & \ding{51} & & \ding{51} & &&&&&&&&
\\
\midrule
\textbf{AdaBrain-Bench} & \ding{51} & \ding{51} & \ding{51} & \ding{51} & \ding{51} & \ding{51} & \ding{51} & \ding{51} & \ding{51} & \ding{51}& \ding{51} & \ding{51} & \ding{51}
\\
\bottomrule
\end{tabular}
}}
\end{table*}

%
%
%
%
%

Despite the advancements in SSP for generalized brain decoding, the field currently suffers from the lack of large-scale and standardized benchmarks for assessing the utility of the brain foundation models in a broad range of BCI tasks.
The development of open benchmarks in natural language and vision domains~\cite{seed-bench, zhang2024benchmarking, tong2024cambrian} have had a transformative impact on the research community.
However, brain decoding lacks comparable evaluation frameworks.
Existing benchmarks~\cite{chevallier2024largest, liu2024libeer} are constrained by small scales, narrow task sets, and limited BCI paradigms. 
For instance, BIOT~\cite{yang2023biot} focuses exclusively on seizure detection, and does not involve other fundamental BCI tasks such as motor imagery and cognitive state assessment. The lack of systematic evaluation restricts our understanding of decoding models' generalizability and practical applicability.
More critically, current evaluation practices for brain foundation models exhibit notable heterogeneity. The evaluated tasks/datasets, evaluation settings, and pipelines vary widely across studies (see Table~\ref{tab:study-comparison}). 
Even for identical tasks, the data splitting strategies and data preprocessing pipelines are inconsistent across studies, causing significant performance fluctuations. 
Therefore, the absence of a standardized benchmark impedes objective methodology comparisons and thorough evaluation of foundation models, limiting both reproducible research and cross-application generalization in BCI systems.  
%

%
%
%
%
%

To tackle these challenges,
this article presents AdaBrain-Bench, a multi-task non-invasive BCI benchmark framework to assess the generalizability of brain foundation models across diverse application scenarios.
AdaBrain-Bench addresses existing benchmark shortfalls by advancing prior work in three distinct ways. 
\textit{Firstly},
the benchmark transcends the boundary of traditional single-task domains by  
incorporating a broad spectrum of representative downstream tasks with varied task paradigms, spanning cognitive state assessment, human augmentation, and clinical monitoring applications.
\textit{Secondly}, AdaBrain-Bench offers a modular pipeline for adapting foundation models to downstream tasks, combining standardized data preprocessing procedures with flexible adaptation strategies. The pipeline not only facilitates seamless and actionable downstream deployment of foundation models in a plug-and-play manner, but also provides flexibility for future expansion of emerging algorithms and tasks. 
\textit{Finally}, the benchmark establishes a multi-faceted evaluation system to rigorously assess the generalization performance of foundation models across diverse real-world application scenarios. The evaluation system defines three distinct settings, namely cross-subject transfer, multi-subject adaptation and few-shot transfer settings, paired with multi-level metrics encompassing both accuracy and transferability metrics to capture key performance indicators of interest.
As a whole, AdaBrain-Bench intends to unify the non-invasive BCI decoding community and brain foundation model research towards a standard benchmark framework, offering a continuously evolving platform to support and accelerate future advancements in the field.

As Figure~\ref{overall} presents, AdaBrain-Bench contains a curated collection of 13 EEG datasets from 7 key BCI tasks in classification, regression, and retrieval task paradigms.
To bridge the pipeline gap and enable direct comparisons, 
the proposed foundation model adaptation pipeline integrates standardized EEG preprocessing, a suite of adaptation strategies, specialized task heads and comprehensive metrics, ensuring broad compatibility across models and datasets.
Beyond conventional accuracy metrics, we introduce a transfer score to assess the improved transferability benefit from pretraining.
Using this pipeline, we evaluate a bunch of baselines in cross-subject transfer, multi-subject adaptation and few-shot transfer settings. These settings target adaptation to multiple unseen subjects, intra-subject variability, and data-limited scenarios, respectively.
We provide a systematic analysis of key factors affecting brain foundation model's generalizability in different scenarios, providing insights for advancing the practice of large-scale EEG pretraining and selecting foundation models.
We offer an live and publicly available benchmark in the official \href{https://github.com/Jamine-W/AdaBrain-Bench}{GitHub repository}.
As new foundation models and datasets are integrated to our benchmark, we will continuously update our findings to provide the community a comprehensive view of progress of foundation models in brain decoding.


%
%





\section{Results}




\subsubsection*{Comprehensive Benchmark for Generalized BCI Decoding}

\begin{table*}[t]
\centering
\caption{\textbf{Benchmarking results in the cross-subject setting.} The brain foundation models are fully fine-tuned to adapt to downstream tasks. The best and second-best results are marked in \textcolor{blue!60}{dark blue} and \textcolor{blue!30}{light blue}, respectively.}
\label{tab:cs}
\setlength{\tabcolsep}{3pt} 
\setlength{\lightrulewidth}{0.2pt} 
\resizebox{1.0\textwidth}{!}{
\begin{tabular}{cc|cccc|cccc}
\toprule
\multirow{2}{*}{\textbf{Dataset}} & \multirow{2}{*}{\textbf{Metrics}} & \multicolumn{4}{c|}{\textbf{Traditional Models}} & \multicolumn{4}{c}{\textbf{EEG Foundation Models}} \\
\cmidrule[\lightrulewidth](lr){3-10} 
 & & \textbf{EEGNet~\cite{Lawhern}} & \textbf{LDMA~\cite{miao2023lmda}} & \textbf{ST-Tran~\cite{Transformer}} & \textbf{Conformer~\cite{Conformer}} & \textbf{BIOT~\cite{yang2023biot}} & \textbf{EEGPT~\cite{eegpt}} & \textbf{LaBraM~\cite{labram}} & \textbf{CBraMod~\cite{wang2024cbramod}} \\
\midrule[0.3pt]
\multirow{2}{*}{\textbf{SEED}}
 & B-Acc & 52.32 & \cellcolor{blue!10}{53.34} & 50.15 & 53.12 & 47.89 & 49.90 & \cellcolor{blue!30}{55.78} & 51.11 \\
 & F1-W & 49.50 & \cellcolor{blue!10}{52.96} & 48.02 & 50.80 & 47.18 & 46.70 & \cellcolor{blue!30}{53.78} & 50.81\\
\arrayrulecolor{mygray} \midrule[0.1pt]
 \multirow{2}{*}{\textbf{SEED-IV}}
 & B-Acc & 34.85 & 36.32 & 32.94 & 34.94 & 35.06 & 31.20 & \cellcolor{blue!30}{40.98} & \cellcolor{blue!10}{39.36} \\
 & F1-W & 28.72 & 35.45 & 33.20 & 33.20 & 33.52 & 29.94 & \cellcolor{blue!30}{40.61} & \cellcolor{blue!10}{38.70} \\
\arrayrulecolor{mygray} \midrule[0.1pt]
 \multirow{2}{*}{\textbf{EEGMAT}}
 & B-Acc & 63.33 & 64.72 & 57.50 & 73.89 & 73.61 & 61.66 & \cellcolor{blue!10}{85.83} & \cellcolor{blue!30}{88.89} \\
 & AUROC & 67.79 & 69.03 & 65.96 & 75.61 & 84.44 & 63.85 & \cellcolor{blue!10}{94.42}& \cellcolor{blue!30}{95.56} \\
\arrayrulecolor{mygray} \midrule[0.1pt]
 \multirow{2}{*}{\textbf{SEED-VIG}}
 & Corr & 58.34 & 57.22 & 49.19 & 52.11 & \cellcolor{blue!10}{62.98} & 57.83 & \cellcolor{blue!30}{65.52} & 60.47 \\
 & R2 & \cellcolor{blue!10}{26.65} & 11.15 & 3.98 & 2.55 & \cellcolor{blue!30}{27.09} & 23.47 & 25.35& 2.40 \\
 \arrayrulecolor{mygray} \midrule[0.1pt]
 \multirow{2}{*}{\textbf{BCI-IV-2A}}
 & B-Acc & \cellcolor{blue!10}{47.83} & 36.20 & 31.42 & 44.88 & 42.53 & 25.81 & \cellcolor{blue!30}{54.98} & 47.71 \\
 & F1-W & 45.46 & 32.55 & 26.84 & 41.54 & 39.70 & 18.52 & \cellcolor{blue!30}{54.90} & \cellcolor{blue!10}{46.25} \\
\arrayrulecolor{mygray} \midrule[0.1pt]
\multirow{2}{*}{\textbf{SHU}}
 & B-Acc & 56.48 & 56.39 & 51.66 & 52.96 & 49.99 & 55.23 & \cellcolor{blue!10}{58.88} & \cellcolor{blue!30}{59.21} \\
 & AUROC & 61.06 & 60.17 & 53.31 & 56.76 & 49.87 & 58.34 & \cellcolor{blue!30}{62.74} & \cellcolor{blue!10}{62.02} \\
\arrayrulecolor{mygray} \midrule[0.1pt]
 \multirow{2}{*}{\textbf{Things-EEG}}
 & 2-Way & 82.58 & 80.93 & \cellcolor{blue!10}{84.40} & 69.92 & 57.90 & 77.30 & \cellcolor{blue!30}{84.50} & 83.20 \\
 & Top-5 & 18.67 & 19.36 & \cellcolor{blue!10}{24.85} & 10.25 & 4.65 & 19.70 & \cellcolor{blue!30}{26.05} & 23.15 \\
\arrayrulecolor{mygray} \midrule[0.1pt]
 \multirow{2}{*}{\textbf{TUEV}}
 & B-Acc & 32.65 & 32.88 & 38.68 & 54.06 & 51.78 & 42.89 & \cellcolor{blue!30}{59.05} & \cellcolor{blue!10}{57.69} \\
 & F1-W & 71.66 & 69.54 & 70.06 & 77.52 & 75.17 & 74.65 & \cellcolor{blue!30}{79.62} & \cellcolor{blue!10}{78.69} \\
\arrayrulecolor{mygray} \midrule[0.1pt]
 \multirow{2}{*}{\textbf{TUAB}}
 & B-Acc & 77.58 & 78.37 & \cellcolor{blue!10}{81.04} & 78.92 & 78.07 & 80.54 & \cellcolor{blue!30}{81.50} & 80.05 \\
 & AUC-PR & 86.48 & 86.97 & \cellcolor{blue!30}{90.41}& 87.95 & 86.93 & 89.36 & \cellcolor{blue!10}{90.08} & 89.19 \\
\arrayrulecolor{mygray} \midrule[0.1pt]
 \multirow{2}{*}{\textbf{Siena}}
 & B-Acc & \cellcolor{blue!10}{72.29} & 66.37 & 64.37 & \cellcolor{blue!30}{72.87} & 71.67 & 59.54 & 66.03 & 65.12 \\
 & AUC-PR & 33.26 & 41.39 & 31.97 & 41.23 & \cellcolor{blue!10}{49.13} & 28.30 & 42.29 & \cellcolor{blue!30}{51.53} \\
\arrayrulecolor{mygray} \midrule[0.1pt]
\multirow{2}{*}{\textbf{HMC}}
 & B-Acc & 66.67 & 70.33 & \cellcolor{blue!10}{73.73} & \cellcolor{blue!30}{73.84} & 70.63 & 70.21 & 71.94 & 71.40 \\
 & F1-W & 67.63 & 76.15 & \cellcolor{blue!30}{77.52} & \cellcolor{blue!10}{77.30} & 74.52 & 74.22 & 74.28 & 72.24 \\
\arrayrulecolor{mygray} \midrule[0.1pt]
\multirow{2}{*}{\textbf{SHHS}}
 & B-Acc & 61.65 & 65.75 & 68.67 & 68.42 & \cellcolor{blue!10}{72.22} & 66.46 & 71.69 & \cellcolor{blue!30}{73.51} \\
 & F1-W & 73.79 & 78.64 & 80.88 & 81.15 & \cellcolor{blue!10}{83.56} & 78.43 & 82.90 & \cellcolor{blue!30}{84.00} \\
\arrayrulecolor{mygray} \midrule[0.1pt]
\multirow{2}{*}{\textbf{Sleep-EDF}}
 & B-Acc & 48.94 & 58.83 & \cellcolor{blue!30}{69.55} & 61.15 & 64.95 & 60.99 & 68.94 & \cellcolor{blue!10}{69.47} \\
 & F1-W & 78.11 & 84.06 & 86.36 & 84.07 & 83.80 & 84.90 & \cellcolor{blue!10}{87.28} & \cellcolor{blue!30}{87.40} \\
\arrayrulecolor{mygray} \midrule[0.1pt]
\multicolumn{2}{c|}{\textbf{Macro-average}} & 56.32 & 56.73 & 55.64 & 58.12 & 58.42 & 55.00 & \cellcolor{blue!30}{64.61} & \cellcolor{blue!10}{62.66} \\
\bottomrule
\end{tabular}
}
\end{table*}


\begin{table*}[tbp]
\centering
\caption{\textbf{Benchmarking results in the multi-subject setting.} The best and second-best results are marked in \textcolor{blue!60}{dark blue} and \textcolor{blue!30}{light blue}, respectively.}
\label{tab:ms}
\renewcommand{\arraystretch}{1.0}
\setlength{\tabcolsep}{4pt} 
\setlength{\lightrulewidth}{0.2pt} 
\resizebox{1.0\textwidth}{!}{
\begin{tabular}{cc|cccc|cccc}
\toprule
\multirow{2}{*}{\textbf{Dataset}} & \multirow{2}{*}{\textbf{Metrics}} & \multicolumn{4}{c|}{\textbf{Traditional Models}} & \multicolumn{4}{c}{\textbf{EEG Foundation Models}} \\
\cmidrule[\lightrulewidth](lr){3-10}
 & & \textbf{EEGNet~\cite{Lawhern}} & \textbf{LDMA~\cite{miao2023lmda}} & \textbf{ST-Tran~\cite{Transformer}} & \textbf{Conformer~\cite{Conformer}} & \textbf{BIOT~\cite{yang2023biot}} & \textbf{EEGPT~\cite{eegpt}} & \textbf{LaBraM~\cite{labram}} & \textbf{CBraMod~\cite{wang2024cbramod}} \\
\midrule
\multirow{2}{*}{\textbf{SEED}}
 & B-Acc & 52.72 & 58.13 & 55.59 & 57.42 & 58.37& 57.50 & \cellcolor{blue!30}{70.90}& \cellcolor{blue!10}{70.31} 
\\
 & F1-W & 53.38 & 58.71 & 55.08 & 57.61 & 58.74 & 58.15 & \cellcolor{blue!30}{71.37} & \cellcolor{blue!10}{70.69}\\
\arrayrulecolor{mygray} \midrule[0.1pt]
 \multirow{2}{*}{\textbf{SEED-IV}}
 & B-Acc & 33.82 & 27.17 & 25.50 & 37.62 & 36.19&  30.57 & \cellcolor{blue!30}{47.63} & \cellcolor{blue!10}{44.20} \\
 & F1-W & 37.25 & 19.42 & 28.71 & 42.13 & 38.67& 28.31 & \cellcolor{blue!30}{49.14} & \cellcolor{blue!10}{45.58} \\
\arrayrulecolor{mygray} \midrule[0.1pt]
 \multirow{2}{*}{\textbf{BCI-IV-2A}}
 & B-Acc & 56.30 & 39.12 & 32.10 & 52.39 & 50.67& 54.01& \cellcolor{blue!30}{60.75}& \cellcolor{blue!10}{59.03}\\ 
 & F1-W & 56.30 & 38.84 & 28.01 & 52.39 & 50.57& 53.86 &  \cellcolor{blue!30}{60.71} & \cellcolor{blue!10}{59.07}\\ 
\arrayrulecolor{mygray} \midrule[0.1pt]
\multirow{2}{*}{\textbf{SHU}}
 & B-Acc & 62.17 & 62.94 & 61.41 & 61.31 & 59.16& 62.87& \cellcolor{blue!30}{67.90}& \cellcolor{blue!10}{66.47} \\ 
 & AUROC & 69.15 & 70.18 & 67.25 & 68.42 & 63.28 & 68.11 & \cellcolor{blue!30}{74.58} & \cellcolor{blue!10}{73.16} \\ 
\arrayrulecolor{mygray} \midrule[0.1pt]
\multicolumn{2}{c|}{\textbf{Macro-average}} & 52.64 & 46.81 & 44.21 & 53.66 & 51.96& 51.67 & \cellcolor{blue!30}{62.87} &  \cellcolor{blue!10}{61.06}  \\
\bottomrule
\end{tabular}}
\end{table*}

We provide a comprehensive benchmark for evaluating brain foundation models' generalization performance on 7 key BCI decoding tasks.
Our benchmark includes diverse datasets and covers critical applications in health monitoring, cognitive monitoring, and human augmentation. The proposed AdaBrain-Bench framework evaluates several recently published EEG foundation models using self-supervised pretraining and traditional non-foundation models.
To ensure robustness of the study results and account for dataset variability, we curated 13 public EEG-based brain decoding datasets with varying electrodes, collected from the task including emotion recognition, workload classification, vigilance estimation, motor imagery classification, visual decoding, clinical anomaly detection, and sleep staging.
For effective adaptation and fair comparison, we established a unified foundation model adaptation pipeline, integrating uniform data splitting, standardized data preprocessing, various task heads and adaptation strategies.
Systematic evaluations were conducted across three distinct evaluation settings, namely cross-subject transfer, multi-subject adaptation and few-shot transfer settings. In addition to conventional accuracy metrics, we utilized transferability metrics for a direct comparison of foundation models' adaptability to downstream tasks. 
In this study, we evaluated several recently published brain foundation models (BIOT~\cite{yang2023biot}, EEGPT~\cite{eegpt}, LaBraM~\cite{labram}, and CBraMod~\cite{wang2024cbramod}). We also selected several representative traditional supervised decoding models (EEGNet~\cite{Lawhern}, LDMA~\cite{miao2023lmda}, ST-Tran~\cite{Transformer}, and Conformer~\cite{Conformer}) for comparison.

\begin{figure}[!htbp]
\centering
\includegraphics[width=0.99\textwidth]{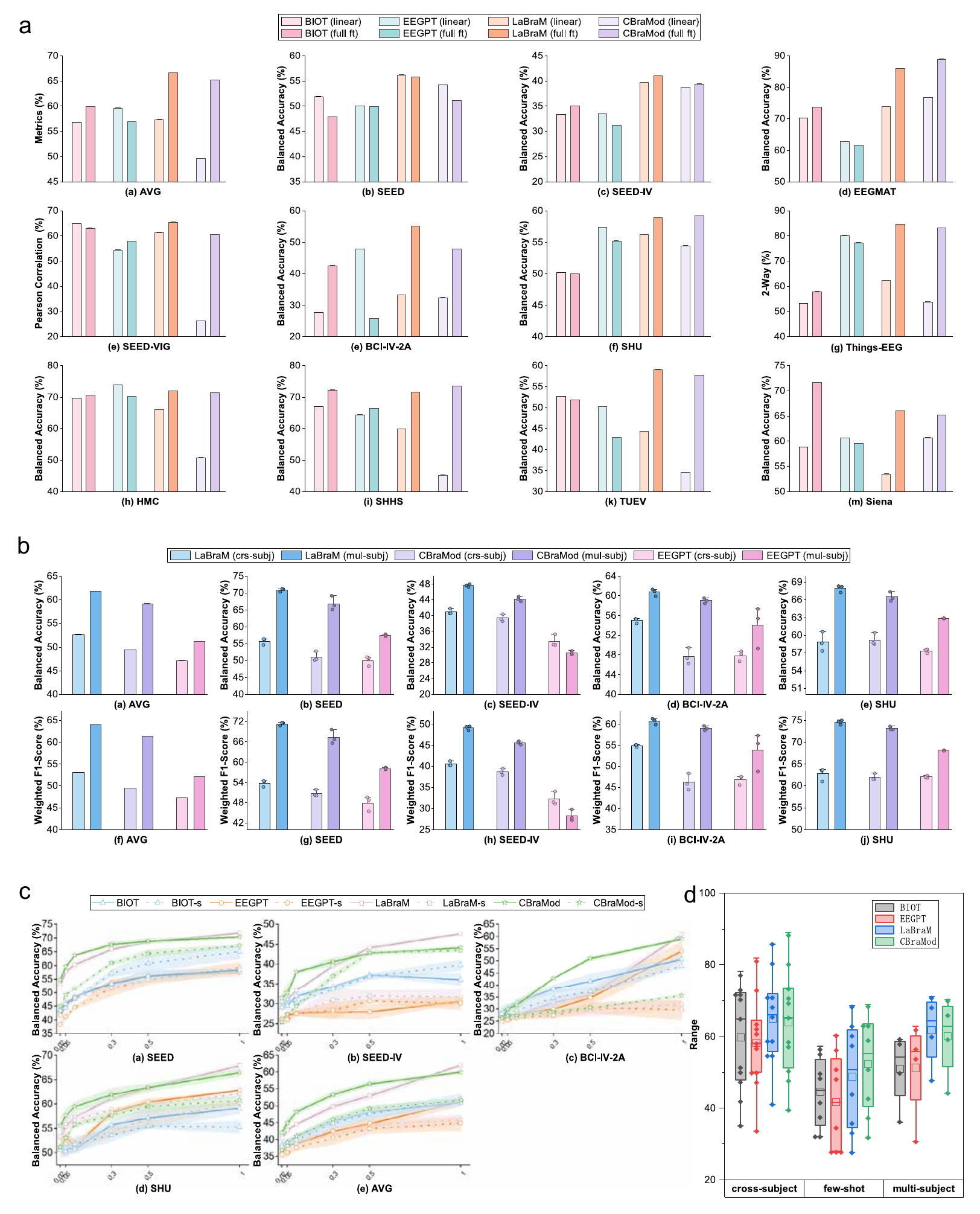}
\caption{\textbf{Performance comparison of brain foundation models in cross-subject, multi-subject, and few-shot settings.}
\textbf{a}, comparison of downstream performance (balanced accuracy) with different fine-tuning strategies: linear probe (linear) and full fine-tuning(full ft). \textbf{b}, comparison of downstream performance (balanced accuracy and weighted F1-score) in multi-subject (mul-subj) vs. cross-subject (crs-subj) settings across all datasets. \textbf{c}, 
benchmarking results in the few-shot setting, with the few-shot sampling ratio $r$ ranging from $[0.02, 0.05, 0.1, 0.3, 0.5]$. Few-shot fine-tuned results of foundation models and their trained-from-scratch counterpart (-s)  are presented.
\textbf{d}, the overall comparison of foundation models across multiple evaluation settings. Each box plots the distribution of evaluated results across all benchmark datasets.
}
\label{4-figures}
\end{figure}


\subsubsection*{Cross-Subject Transfer} 
Substantial inter-subject variability in EEG signals presents a critical challenge for developing generalizable decoding models. In this study, we introduce a cross-subject transfer setting where foundation models are finetuned on a subset of subjects and required to simultaneously generalize to a distinct cohort of previously unseen subjects, testing model's capacity to capture subject-shared general neural patterns.
Table~\ref{tab:cs} reports the benchmarking results for brain foundation models and traditional supervised models in the cross-subject transfer setting, with foundation models fully fine-tuned. 
%
We observe that foundation models consistently outperform traditional baselines trained from scratch across metrics on most of datasets, claiming the best or second best results on the majority of tasks. 
For instance, for EEGMAT (workload classification), foundation models claim high performance of over 80\% in B-Acc (LaBraM 85.83\% and CBraMod 88.89\%, respectively), while the best traditional model obtains accuracy of 73.89\%. For SHHS, the SOTA traditional approach had an accuracy of 68.67\%, while BIOT and CBraMod achieved higher accuracies of 72.22\% and 73.51\%, respectively. 
%
However, for clinical monitoring tasks, including TUAB, Siena, HMC, and Sleep-EDF, foundation models perform comparable or even worse than traditional models. For instance, on Sleep-EDF, the best foundation model result is 69.47\%,  compared to 69.55\% for the best traditional model result.
This may be attributed to the large volume of EEG recordings in these clinical monitoring datasets, which lowers the generalization complexity and thus reduces the performance gap.
Overall, the macro-average of primary and secondary metrics show that well-trained foundation models significantly outperform traditional models, with LaBraM and CBraMod achieving scores of 64.61\% and 62.66\%, respectively, compared to 58.12\% for the SOTA traditional model. The results demonstrate that brain foundation models show great potential for enhancing cross-subject generalization performance.

Among models, LaBraM and CBraMod consistently achieved higher decoding performance than other models, achieving the best and second-best Macro-average scores. 
We hypothesize that this is because these models are pretrained on a larger amount of data and have better channel adaptability by learning adaptive spatial and temporal position embeddings. 
%
Individual tasks exhibited considerable potential for improvement in the cross-subject transfer setting. Some tasks like emotion recognition (\textit{i.e.}, SEED, SEED-IV) and motor imagery (\textit{i.e.}, BCI, SHU) yielded suboptimal performance across methods. This can be explained by the greater inherent challenges of these tasks due to high individual variability in mental states.

Apart from using full fine-tuning strategy for adaptation, we also considered the linear probe strategy that fine-tunes an additional task head while keeping the whole pretrained models frozen, offering a computationally efficient adaptation approach. 
Figure~\ref{4-figures}a presents benchmarking results in the cross-subject transfer setting using linear probe, compared with the full fine-tuning results.
%
Results demonstrate that linear probing remains a challenging setting. Compared to fine-tuning, linear evaluation consistently yields lower performance across most datasets, including those with limited samples (\textit{e.g.}, EEGMAT). For instance, for BCI-IV-2A, CbraMod attains 47.71\% accuracy with fine-tuning but only 32.32\% with linear probing.
%
It is worth noting that EEGPT demonstrates stronger performance using linear probing compared to full fine-tuning across certain datasets. For instance, on BCI-IV-2A and HMC, the linear probing performance is 47.89\% and 73.82\%, respectively, compared to 25.81\% and 70.21\% with full fine-tuning performance.
It can be attributed to EEGPT's large parameter count (see Table~\ref{tab:summary-model}) that increases the risk of overfitting when using the full fine-tuning strategy. 
We can infer that for large-scale foundation model, linear probing may better preserve their brainwave representation capabilities. 
%

\subsubsection*{Multi-Subject Adaptation}
To evaluate the adaptability of models against intra-subject variability, we introduce a multi-subject adaptation setting wherein models are trained on EEG data from multiple subjects and subsequently tested on data from distinct recording trials or sessions of the same subject cohort.
We follow established methodologies~\cite{labram, wang2024cbramod, yang2023biot} that employ similar within-subject setting to assess model  performance on emotion recognition task (\textit{i.e.}, SEED, SEED-IV) and motor imagery classification task (\textit{i.e.}, BCI-IV-2A, SHU). 
The non-stationarity of EEG signals poses significant challenges for brain decoding. This is due to variations in factors like electrode placement, the participant's physiological state and environmental conditions over different recording sessions or trials.

%
Table~\ref{tab:ms} reports the decoding performance for brain foundation models and traditional models in the multi-subject adaptation setting. Hereafter we report linear probing results for EEGPT given its better performance, while employing full fine-tuning for other foundation models.
The results demonstrate a more consistent trend across methods compared to the cross-subject transfer setting. 
We observe notable differences in the decoding performance that emerged among the foundation models.
LaBraM and CBraMod achieved the top-2 macro-average scores, consistently outperforming traditional decoding methods across all datasets with an average score advantage exceeding 6 points. For instance, on SEED, LaBraM and CBraMod attained accuracies of 70.90\% and 70.31\%, respectively, significantly exceeding the SOTA traditional model (LDMA 58.13\%). 
In contrast, the multi-subject performance of other foundation models showed a performance gap compared to traditional models. BIOT and EEGPT achieved macro-average scores of 51.96\% and 51.67\%, respectively, falling below traditional models such as EEGNet (52.64\%) and Conformer (53.66\%). 


\begin{figure}[tbp]
\centering
\includegraphics[width=0.99\textwidth]{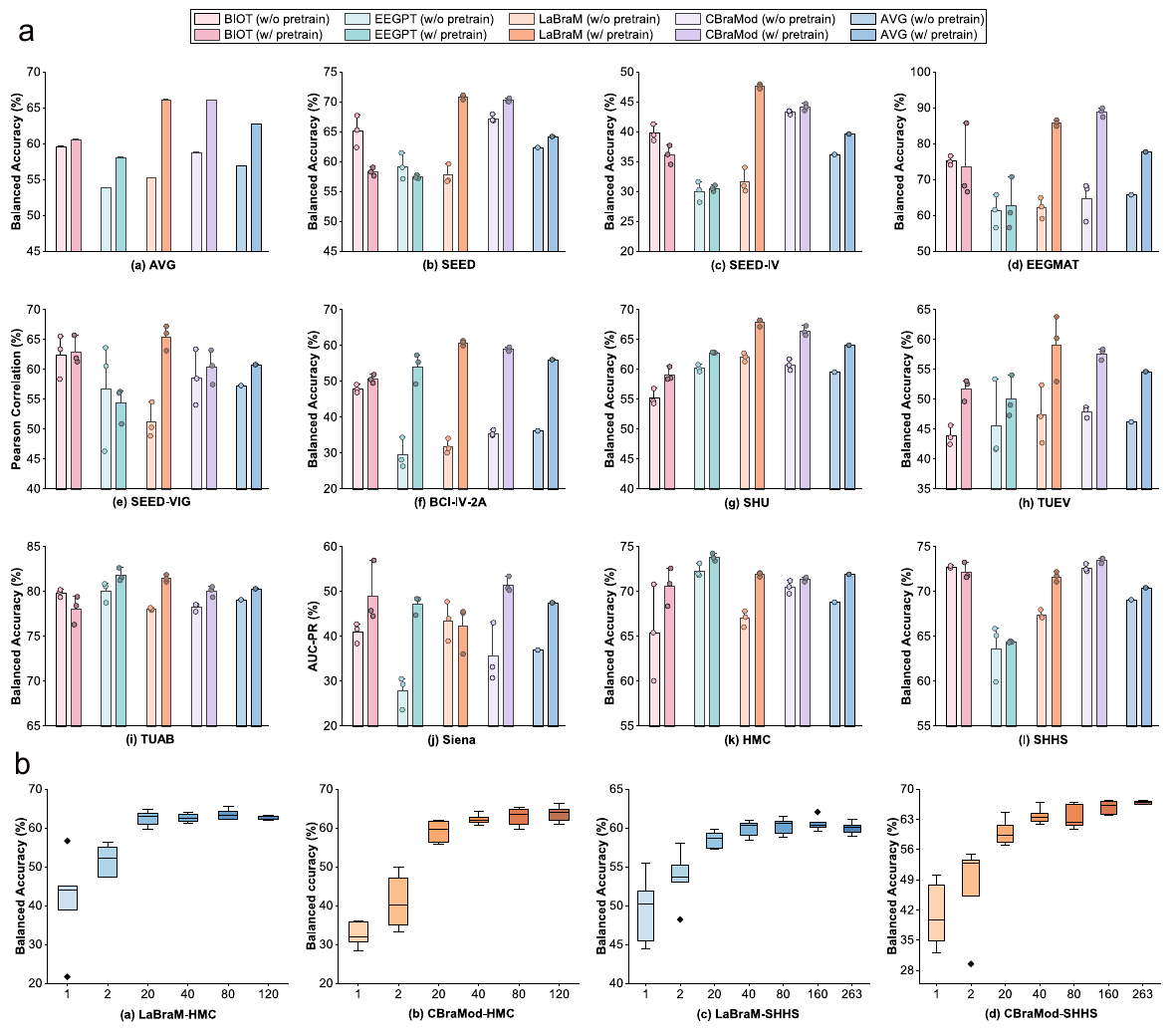}
\caption{\textbf{Impacting factors vs. downstream performance.}
\textbf{a}, comparison of downstream performance (balanced accuracy) with vs. without the pre-training stage across multi-subject (mul-subj) and cross-subject (crs-subj) settings. 
\textbf{b}, the downstream task performance shows a positive correlation with the number of subjects in the training set in the cross-subject setting.
}
\label{2-figures}
\end{figure}

\noindent \textbf{Comparison between Cross-Subject and Multi-Subject Settings.} 
We further conduct a cross-setting comparison analysis between multi-subject and cross-subject results.
As presented in Figure~\ref{4-figures}b, the cross-subject results are consistently lower than multi-subject results across models.
For example, on SEED, the cross-subject performance of LaBraM, CBraMod, and EEGPT are lower than multi-subject performance by 15\%, 19\%, and 7\%, respectively. 
Foundation models demonstrated superior generalization when transferring across sessions than across subjects. 
These results hint that inter-subject variability poses a greater challenge for decoding models than the inter-trial or session variability for the same subjects. 
Individual variations in brain anatomy, functional organization and cognitive experience lead to highly diverse EEG signal characteristics, making it exceptionally difficult to develop foundation models with subject-universal representation capacity. 
The multi-subject setting could serve as a crucial baseline for generalization research, acting as a natural upper bound to quantify the performance drop when attempting to generalize across subjects.

\subsubsection*{Impact of Pretraining}

To explore how self-supervised pretraining strategies impact the performance, we directly trained foundation models from scratch on downstream datasets without any pretraining across multi-subject and cross-subject settings, as shown in Figure~\ref{2-figures}a.
%
We can observe that the utilization of pretraining strategies brings substantial improvement across different foundation methods. LaBraM and CBraMod with pretraining outperform the non-pretrained counterpart by over 10\% and 7\% in average, respectively. 
Based on these results, we highlight that large-scale self-supervision through  masked signal modeling can effectively discover functionally meaningful brain patterns from unlabeled EEG data, improving transferability to different BCI tasks. 
Notably, the performance gains are typically more pronounced on datasets with limited scale of training data, including BCI-IV-2A and EEGMAT, compared to datasets with large amount of data, including SHHS, HMC, and TUAB. 
These results hinted that pretrained foundation models may be particularly advantageous for data-limited datasets. 
However, pretraining strategies show limited effectiveness in certain cases.
An observation is that the performance improvement from EEGPT and BIOT pretraining is less pronounced compared to other methods, with limited improvement on some datasets.
This likely stems from EEGPT's relatively small pretraining data amount (246 hours of EEG data), compared to LaBraM (2,500 hours) and CBraMod (27,000 hours). The constrained pretraining dataset size may cause insufficient learning of universal neural representations, particularly given its significantly larger model architecture (25M vs. 5.8M/4M). This suggests model scaling should be proportional to available training data. 
For BIOT, the bottleneck likely stems from its support for fixed 18 channels in its pretrained version, which is incompatible with downstream tasks such as SEED and SEED-IV (62 channels), potentially diminishing the spatial cues of downstream EEG signals.

\subsubsection*{Few-Shot Transfer}
To investigate the models' generalization capacity in the data-limited scenario, we design a few-shot transfer setting where the models are fine-tuned with a small set of training data from the downstream task. 
For each dataset, a subset of the training data is used as the support set for training, with the sampling ratios $r$ ranging from $[0.02, 0.05, 0.1, 0.3, 0.5]$. Figure~\ref{4-figures}c presents the decoding performance in the few-shot transfer setting, with the full-data performance (\textit{i.e.}, $r=1$) also presented.
The few-shot performance consistently lags behind the full-data performance by a significant margin, especially for emotion recognition tasks (SEED and SEED-IV). The accuracies exhibit a clear upward trend as the amount of training data increases.
For motor imagery classification datasets (SHU and BCI-IV-2A), the accuracy rapidly grows with just a few dozen additional samples. For emotion recognition datasets (SEED, SEED-IV), the trend of performance improvement gradually slows down. On SEED-IV, EEGPT’s performance is significantly lower than other methods across few-shot ratios, achieving an accuracy of 27.93\% at $r=0.5$ compared to BIOT's 37.36\%, indicating potential overfitting in high-parameter models.
%
We observe that the foundation model's performance lead compared to models without pretraining is pronounced in the data-limited settings. On SEED, CBraMod outperforms non-pretrained version by 10\%, {12\%} when $r=[0.05, 0.1]$, and by {3\%} when using the full set of data.
On BCI-IV-2A, CBraMod achieves gains of {14.09\%} when using 33\% training data compared to models trained from scratch. These findings suggest that the employment of self-supervised pretraining could enhance the generalizability of foundation models when only limited EEG recordings are available.

%


%

\subsubsection*{Quantitative Assessment of the Transferability}
We have previously hinted at the improved transferability of the foundation model via the use of a self-supervised strategy.
We further quantify the enhanced transferability of the foundation models by introducing the \textit{\textbf{transfer score} (TS)}, which is defined as the relative improvement of the pretrained foundation model over its trained-from-scratch counterpart (see Methods Section for more details). Table~\ref{tab:ts} presents the TS in the cross-subject transfer and few-shot transfer settings. Self-supervised foundation models show positive transferability in most cases and exhibit distinct advantages in different settings. 
CBraMod exhibits superior transfer scores under the few-shot transfer setting. This may be attributed to its adoption of input-conditioned channel embedding that adapts flexibly to datasets with varying channel configurations, a challenge that becomes more pronounced under few-shot constraints.

\begin{table*}[tbp]
\centering
\caption{\textbf{The transfer score in various settings.} The table presents the TS in the cross-subject transfer setting (Cross-subject TS) and the few-shot transfer setting (Few-shot TS) with different few-shot sampling ratios. The \textbf{bold} denotes the best results.}
\label{tab:ts}
\renewcommand{\arraystretch}{1.2}
\setlength{\tabcolsep}{16pt}
\resizebox{1.0\textwidth}{!}{
\begin{tabular}{c|c|ccccc}
\toprule
\multirow{2}{*}{\textbf{Method}} & \multirow{2}{*}{\textbf{Cross-subject TS}} & \multicolumn{5}{c}{\textbf{Few-shot TS}}
\\ \cmidrule{3-7}
& & \textbf{0.02} & \textbf{0.05} & \textbf{0.1} & \textbf{0.3} & \textbf{Average} \\
\midrule
\textbf{BIOT}~\cite{yang2023biot} & 0.0264 & 0.0402 & 0.0111 & 0.0683 & -0.3489      & -0.0573 \\
\textbf{EEGPT}~\cite{eegpt} & 0.0764 & 0.0209 & 0.1159 & 0.0651 & 0.1135 & 0.0789 \\
\textbf{LaBraM}~\cite{labram}  & \textbf{0.2373} & 0.1092 & 0.1231 & 0.1783 & 0.2987 & 0.1773 \\
\textbf{CBraMod}~\cite{wang2024cbramod} & 0.1391 & \textbf{0.1937} & \textbf{0.2315} & \textbf{0.3096} & \textbf{0.3695} & \textbf{0.2761}   
\\
\bottomrule
\end{tabular}}
\end{table*}

\subsubsection*{Impact of Number of Training Subjects}
Given the challenge of cross-subject transfer setting, how to increase cross-subject performance is a critical issue. We thus assess how the number of subjects used in training can impact the model's ability to generalize to new subjects. We fine-tune LaBraM and CbraMod with different numbers of subjects, denoted as $N_{subj}$, in the cross-subject transfer setting. Datasets with a large amount of subjects were selected for this experiment, including HMC and SHHS. These two datasets are both sleep staging datasets, including 120 and 263 subjects, respectively. To isolate the effect of sample amount on the performance, we keep identical sample counts per epoch for each dataset. The results with varying $N_{subj}$ are reported in Figure~\ref{2-figures}b.
We observe a steady improvement in cross-subject decoding accuracies on unseen subjects as the number of training subjects increases in the early phase. When trained on a limited number of subjects, model performance is consistently unsatisfactory across methods and datasets compared to training with all subjects. For instance, on HMC, utilizing only two subjects leads to {10.75\%} and {18.19\%} lower performance for LaBraM and CbraMod, respectively, compared to using 20 subjects. On SHHS, CbraMod attains a balanced accuracy of {40.70\%} with a single subject, while demonstrating a substantial {23.09\%} increase in performance when 40 subjects are used for training. Performance tends to plateau at approximately 40 subjects on both HMC and SHHS. After $N_{subj}=40$, further increasing training subjects provides only marginal gains. These findings suggest that expanding the training cohort size may be an effective way to enhance cross-subject generalization performance.

\subsubsection*{Impact of Normalization}
Due to the inherent non-stationarity and high noises in EEG signals, the signal normalization techniques may have a non-negligible influence on decoding. The distribution gap between pretrained datasets and downstream tasks may further exacerbate this impact. Various normalization choices have been utilized in previous task-specific methods, such as z-score~\cite{eegpt},  unit rescale~\cite{labram}, and 95-percentile normalization~\cite{yang2023biot}. How to select the optimal normalization strategy becomes a critical problem when adapting foundation models to downstream tasks.
Thus, we studied the impact of different normalization choices, including \textbf{z-score}, \textbf{95-percentile}, and \textbf{unit rescale} normalization. 
Z-score normalization standardizes the EEG data $X$ by subtracting the mean ($\mu$) and dividing it by the standard deviation ($\sigma$), resulting in a distribution with zero mean and unit variance. 95-percentile normalization uses the 95-percentile of the absolute amplitude to normalize each channel. Unit rescale normalization directly rescales the unit magnitude of the signal to ensure most values fall within the range of approximately [-1,1].
As reported in Table~\ref{tab:normalization},
our results indicate that z-score turns out to be a robust normalization strategy, generally yielding better performance across the majority of datasets. Z-score  ensures scale invariance and demonstrates an advantage in handling non-stationary processes. However, in datasets exhibiting substantial outliers (\textit{e.g.}, SHU), 95-percentile normalization demonstrates better performance. 
This rank-based normalization enhances robustness to outliers by reducing reliance on extreme values, making it suitable for datasets containing such anomalies.


\begin{table*}[tbp]
\centering
\caption{\textbf{Decoding performance across various normalization techniques.}  We studied the impact of three commonly-used normalization choices, including \textbf{z-score}, \textbf{95-percentile}, and \textbf{unit rescale} normalization. 
For SEED, SEED-IV, BCI-IV-2A, and SHU, we followed the multi-subject setting for evaluation. The remaining datasets were evaluated using cross-subject setting. BIOT, LaBraM and CBraMod are evaluated under full fine-tuning setting. The best performance for each model is highlighted in \textbf{bold}.}
\label{tab:normalization}
\resizebox{1.0\textwidth}{!}{
\begin{tabular}{cc|cccccccccc}
\toprule
\multirow{2}{*}{\textbf{Dataset}} & \multirow{2}{*}{\textbf{Metrics}} & \multicolumn{3}{c}{\textbf{BIOT}~\cite{yang2023biot}} & \multicolumn{3}{c}{\textbf{LaBraM}~\cite{labram}} & \multicolumn{3}{c}{\textbf{CBraMod}~\cite{wang2024cbramod}} \\
\cmidrule(lr){3-5} \cmidrule(lr){6-8} \cmidrule(lr){9-11}
& & {Unit Rescale} & {95-percentile} & {Z-Score} & {Unit Rescale} & {95-percentile} & {Z-Score} & {Unit Rescale} & {95-percentile} & {Z-Score} \\
\midrule
\multirow{2}{*}{SEED} & B-Acc & 57.59 & \textbf{59.31} & 58.37 & 70.94 & \textbf{71.91} & 70.90 & 65.95 & 67.33 & \textbf{70.31} \\
& F1-W & 57.79 & \textbf{59.79} & 58.74 & 71.22 & \textbf{72.26} & 71.37 & 66.41 & 67.82 & \textbf{70.69} \\
\midrule
\multirow{2}{*}{SEED-IV} & B-Acc & 35.61 & 33.72 & \textbf{36.19} & 46.80 & \textbf{47.92} & 47.63 & 39.87 & 36.92 & \textbf{44.20} \\
& F1-W & 35.14 & 34.12 & \textbf{38.67} & 48.59 & \textbf{49.23} & 49.14 & 36.89 & 37.33 & \textbf{45.58} \\
\midrule
\multirow{2}{*}{EDMAT} & B-Acc & 72.50 & 70.55 & \textbf{73.61} & \textbf{85.83} & 77.50 & \textbf{85.83} & 76.94 & 75.83 & \textbf{88.89} \\
& AUROC & 83.57 & 75.62 & \textbf{84.44} & 94.32 & 86.26 & \textbf{94.42} & 86.29 & 83.95 & \textbf{95.56} \\
\midrule
\multirow{2}{*}{SEED-VIG} & Corr & 52.27 & 52.59 & \textbf{62.98} & 64.00 & \textbf{69.09} & 65.52 & 47.51 & 51.43 & \textbf{60.47} \\
& R2 & 17.67 & 8.65 & \textbf{27.09} & 10.86 & \textbf{41.47} & 25.35 & -17.19 & \textbf{4.29} & 2.40 \\
\midrule
\multirow{2}{*}{BCI-IV-2A} & B-Acc & 50.05 & 48.89 & \textbf{50.67} & 60.24 & 57.38 & \textbf{60.75} & 57.46 & 56.38 & \textbf{59.03} \\
& F1-W & 49.92 & 48.82 & \textbf{50.57} & 60.19 & 57.35 & \textbf{60.71} & 57.41 & 56.37 & \textbf{59.07} \\
\midrule
\multirow{2}{*}{SHU} & B-Acc & 54.70 & \textbf{59.16} & 50.36 & 66.01 & \textbf{67.90}& 65.25 & 64.24 & \textbf{66.47} & 64.12 \\
& AUROC & 56.62 & \textbf{63.28} & 49.61 & 72.22 & \textbf{74.58} & 71.50 & 70.53 & \textbf{73.16} & 70.42  \\
\midrule
\multirow{2}{*}{TUEV} & B-Acc & 50.31 & 47.12 & \textbf{51.78} & \textbf{60.33} & 56.41 & 59.05 & \textbf{58.50} & 56.38 & 57.69 \\
& F1-W & \textbf{76.57} & 74.85 & 75.17 & \textbf{81.35} & 78.16 & 79.62 & \textbf{79.06} & 77.62 & 78.69  \\
\midrule
\multirow{2}{*}{TUAB} & B-Acc & 78.03 & \textbf{78.94} & 78.07 & 81.29 & 80.32 & \textbf{81.50} & \textbf{80.45} & 79.79 & 80.05 \\
& AUC-PR & 86.77 & \textbf{87.85} & 86.93 & 89.64 & 88.92 & \textbf{90.08} & \textbf{89.39} & 88.68 & 89.19  \\
\midrule
\multirow{2}{*}{Siena} & B-Acc & 62.53 & 60.31 & \textbf{71.67} & 57.36 & 56.36 & \textbf{66.03} & \textbf{69.75} & 52.07 & 65.12  \\
& AUC-PR & 38.16 & 25.64 & \textbf{49.13} & \textbf{46.54} & 38.95 & 42.29 & \textbf{51.81} & 26.57 & 51.53 \\
\midrule
\multirow{2}{*}{HMC} & B-Acc & 71.78 & \textbf{72.57} & 70.63 & 71.43 & 71.83 & \textbf{71.94} & 71.72 & \textbf{73.27} & 71.40 \\
& F1-W & 75.75 & \textbf{76.28} & 74.52 & 73.57 & 72.25 & \textbf{74.28} & 72.90 & \textbf{74.02} & 72.24 \\
\midrule
\multirow{2}{*}{SHHS} & B-Acc & 72.14 & 71.87 & \textbf{72.22} & 71.57 & 69.87 & \textbf{71.69} & \textbf{73.86} & 69.93 & 73.51  \\
& F1-W & 83.34 & 83.01 & \textbf{83.56} & 81.87 & 79.19 & \textbf{82.90} & 83.71& 80.09 & \textbf{84.00}  \\
\midrule
\multirow{2}{*}{Sleep-EDF} & B-Acc & 63.59 & 56.55 & \textbf{64.95} & 67.87 & 68.50 &\textbf{68.96} & 69.09 & \textbf{69.53} & 69.47 \\
& F1-W & 83.33 & 83.88 & \textbf{83.80} & 86.54 & 87.27 & \textbf{87.29} & 85.84 & 87.39 & \textbf{87.40} \\
\bottomrule
\end{tabular}
}
\end{table*}

\section{Discussions}

Self-supervised pretrained foundation models have the potential to revolutionize brain decoding techniques for non-invasive BCI systems. 
Brain foundation models trained on large-scale EEG recordings are showing a clear benefit compared to traditional supervised models in terms of generalizability in various settings. While there is still much work to be done towards adapting decoding models to downstream real-world BCI systems, the emergence of foundation models will likely become pivotal to future advances.
As more and more foundation models are trained, a comprehensive benchmark including wide-ranging tasks and a unified evaluation pipeline becomes essential for both researchers training foundation models and looking to apply these pretrained foundation models on downstream tasks. Developing new foundation models requires substantial resources, making it essential to build upon previous efforts. A well-designed benchmark can provide insights for improving pretraining and models in the future. 
For downstream BCI applications, such benchmarks can guide the model selection by systematically evaluating performance across various tasks and experimental conditions.
In this work, we presented AdaBrain-Bench, a comprehensive benchmark focusing on 13 brain decoding datasets. We provided a benchmark study of publicly available brain foundation models to investigate their generalization performance in cross-subject transfer, multi-subject adaptation, and few-shot transfer settings. Importantly, the evaluated data covers a wide range of BCI applications, spanning cognitive state assessment, human augmentation, and clinical monitoring. The benchmark code and instructions are provided in \href{https://github.com/Jamine-W/AdaBrain-Bench}{GitHub repository}.

Pre-trained brain foundation models show significant potential to improve cross-subject generalization across BCI applications.
LaBraM~\cite{labram} and CBraMod~\cite{wang2024cbramod} demonstrate good transferability across various datasets and BCI tasks (see Table~\ref{tab:cs}), which can be explained by their self-supervised pretraining that can effectively extract meaningful features from large, heterogeneous EEG data. As evidenced in Figure~\ref{2-figures}a, the adoption of pretraining strategies consistently enhances performance across diverse foundation models and datasets.
Besides large-scale pretraining, another reason could be their better channel compatibility enabled by flexible channel coverage and adaptive spatial-temporal position embeddings. 
EEG datasets present significant channel variability, and the gap between pretrained channels and downstream channels may lead to neural pattern missing. 
Overall, large-scale and diverse pretraining data sources and high compatibility with heterogeneous EEG configurations are crucial for strong foundation models.
%
Despite the improvement, individual tasks still have significant room for cross-subject generalization improvement. On some tasks like emotion recognition (\textit{i.e.}, SEED, SEED-IV) and motor imagery (BCI, SHU),
foundation models exhibit suboptimal performance and require further improvement before being available for real-world usage. 
We hypothesize that cross-subject generalization challenges stem from highly individual neural responses shaped by personal experiences and mental imagery in these tasks, which introduce inherent subjectivity and complicate the identification of neural patterns across subjects.  
%
%
In contrast, clinical monitoring tasks like sleep stage classification and clinical anomaly detection show comparably higher performance, as they present more consistent cross-subject EEG patterns that serve as good biological markers. 
Consequently, there is an urgent need to develop better cross-subject decoding models, either to improve pretraining strategy to learn subject-shared patterns, or to develop efficient learning strategies for fast cross-subject adaptation. 
As an initial attempt, we investigated whether the subject number could impact the cross-subject performance. 
The results in Figure~\ref{2-figures}b suggest that expanding the training cohort size may effectively improve the adaptability of foundation models across subjects.

In the multi-subject adaptation setting, the performance of certain foundation models also generally surpasses traditional models. 
This underscores the good adaptability of foundation models in handling EEG's inherent non-stationarities of the same subject cohort.
We also conduct cross-setting comparison analysis. As shown in Figure~\ref{4-figures}b, the cross-subject results consistently lag behind multi-subject results, which further confirms the generalization difficulty in cross-subject transfer setting. 
This indicates that inter-subject variability presents a greater challenge for foundation models than inter-trial or session variability of the same subject cohort.

Compared to multi-subject and cross-subject settings, few-shot transfer presents a greater generalization challenge.
Current models show substantial performance gaps between few-shot and full-data conditions, as shown in Figure~\ref{4-figures}c, underscoring the need for foundation models with more robust and generalizable neural representations.
%
We can also observe significantly better performance from foundation models compared to their non-pretrained counterparts in few-shot setting. This advantage likely stems from their ability to transfer knowledge acquired during large-scale pretraining to novel tasks with limited data.
We further quantify models' transferability via the transfer score over the non-pretrained counterpart across settings, as shown in Table~\ref{tab:ts}. Results suggest that large-scale self-supervised pretraining, when combined with flexible handling of channel variability, could consistently enhance the transferability of neural decoding models in both cross-subject and few-shot transfer settings.

We systematically investigated key factors affecting the downstream task performance.
Regarding adaptation strategies, the results in Figure~\ref{4-figures}a reveal that full fine-tuning yields superior performance for most models compared to linear probing. While the latter offers stronger computational efficiency, the current foundation models are not prepared for direct generalization with frozen representations. 
While self-supervised learning captures meaningful representations, they often lack linear separability without full fine-tuning. 
A promising solution could be extensive pretraining with combined model and data scaling to learn more generalizable neural representations. 
For large-scale models like EEGPT, linear probing proves to be a more effective adaptation strategy, as it mitigates the risk of over-fitting of when fine-tuning on limited EEG data.
Regarding the normalization techniques, our results (see Table~\ref{tab:normalization}) indicate that z-score normalization is the most robust preprocessing strategy for aligning downstream data with foundation model representations. 


This study has several limitations that should be acknowledged. Due to the rapidly evolving nature of this field, our study cannot cover all tasks and models. Instead, we focus on establishing baseline performance for key BCI applications using representative models, while ensuring full transparency through publicly released data, evaluation pipelines, and code. This enables downstream users to fairly benchmark new foundation models and associated methods, helping them to determine the applicability of foundation models in their specific use cases.  Future work will prioritize inclusivity by expanding the scope to incorporate a wider range of foundation models and BCI tasks.



\section{Methods}

\subsection*{Benchmark Tasks}

To comprehensively evaluate the EEG foundation models, AdaBrain-Bench selects 7 popular tasks in EEG-based BCIs, covering critical applications in cognitive state assessment (\textit{i.e.}, emotion recognition, workload classification, vigilance estimation), human augmentation (\textit{i.e.} motor imagery classification, visual decoding), and clinical monitoring (\textit{i.e.}, clinical anomaly detection, sleep staging), ranging from passive BCIs to active BCIs. The following sections provide a detailed introduction of involved tasks and datasets. Table~\ref{tab:bci-datasets} summarizes the downstream BCI tasks and datasets currently included in AdaBrain-Bench. 


\noindent \textbf{Emotion Recognition} aims at decoding human emotional experiences by recognizing human emotional states via EEG analysis, enabling early detection of emotional disorders. This task leverages the \textbf{SEED}~\cite{zheng2015investigating} and \textbf{SEED-IV}~\cite{zheng2018emotionmeter} datasets, which both compromise EEG recordings collected from 15 participants exposed to emotion-evoking video stimuli. 
The SEED dataset classifies emotional responses into three categories: negative, neutral, and positive. The SEED-IV dataset provides a more granular classification, categorizing EEG signals into neutral, sad, fearful, and happy states. Both datasets contain EEG recordings acquired from 62 channels.

\noindent \textbf{Workload Classification} aims to assess cognitive load by analyzing EEG signals collected from subjects under varying mental demands, facilitating
real-world applications in cognitive training and early detection of cognitive fatigue in high-stakes environments.
This task is based on \textbf{EEGMAT} dataset~\cite{zyma2019electroencephalograms, goldberger2000physiobank}, which comprises 19-channel EEG recordings collected from 36 participants engaged in mental arithmetic tasks. 
The dataset provides binary labels that distinguish between resting states (low-load) and mental arithmetic working states (high-load).


\noindent \textbf{Vigilance Estimation} aims to measure fatigue levels from EEG signals, enabling real-time monitoring of individuals in high-risk environments and helping prevent fatigue-related accidents. 
This task is based on \textbf{SEED-VIG} dataset~\cite{SEED-VIG}. 
21 subjects were instructed to perform simulated driving tasks, during which they gradually transitioned from vigilance to fatigue. EEG signals were recorded from 17 channels throughout the task, and fatigue levels were annotated using the PERCLOS indicator obtained from SMI eye-tracking glasses.


%
\noindent \textbf{Motor Imagery Classification} aims at decoding the intended motor movements by analyzing brain activity from EEG signals during the motor imagery process, where the decoded signals can be leveraged for controlling external devices for individuals with movement disorders.
This task leverages the BCI Competition IV-2A(\textbf{BCI-IV-2A}) dataset~\cite{tangermann2012review} and \textbf{SHU} dataset~\cite{ma2022large}.
BCI-IV-2A contains 22-channel EEG signals collected from 9 subjects, capturing $4$ classes of motor imagery: left hand, right hand, feet, and tongue. 
SHU dataset consists of 32-channel EEG signals from 25 subjects, including two motor imagery tasks: left hand and right hand.

\noindent \textbf{Visual Decoding} aims at decoding the visual experience that individuals perceived or imagined from brain activity~\cite{NEURIPS2024_ba5f1233}. 
For this task, we utilize the Things-EEG dataset~\cite{gifford2022large, grootswagers2022human}, which comprises 63-channel EEG recordings from $10$ subjects viewing $16,540$ images across various object categories. 
We follow previous work~\cite{ATM} to adopt an EEG-based image retrieval task to evaluate the decoding performance.
%

\begin{table*}[tbp]
\centering
\caption{\textbf{Summary of downstream BCI tasks and datasets currently included in AdaBrain-Bench.} For each dataset, we present its primary and secondary accuracy metrics.}
\label{tab:bci-datasets}
\renewcommand{\arraystretch}{1.0}
\setlength{\tabcolsep}{5pt}
\resizebox{0.99\textwidth}{!}{
\begin{tabular}{cccccccccc}
\toprule
\textbf{\makecell{Application\\Domain}} & \textbf{Task} & \textbf{Dataset} & \textbf{\makecell{Sampling\\Rate}} & \textbf{\#Channel} & \textbf{Duration} & \textbf{\#Subj} & \textbf{\#Samples} & \textbf{\makecell{Primary\\Metric}} & \textbf{\makecell{Secondary\\Metric}}\\
\midrule
\multirow{6}{*}{\textbf{\makecell{Cognitive State \\ Assessment}}} 
& \multirow{2}{*}{\makecell{Emotion\\Recognition}}
& SEED~\cite{zheng2015investigating} & 1,000Hz & 62 & 1s & 15 & 144,852 & B-Acc & F1-W\\
& & SEED-IV~\cite{zheng2018emotionmeter} & 1,000Hz & 62 & 1s & 15 & 151,845 & B-Acc & F1-W\\
\cmidrule{2-10}
& \makecell{Workload \\Classification}
& EEGMAT~\cite{zyma2019electroencephalograms, goldberger2000physiobank} & 500Hz & 19 & 4s & 36 & 1,080 & B-Acc & AUROC\\
\cmidrule{2-10}
& \makecell{Vigilance\\Estimation}
& SEED-VIG~\cite{SEED-VIG} & 200Hz & 17 & 8s & 21 & 20,355 & Corr & R2\\
\midrule
\multirow{4}{*}{\textbf{\makecell{Human \\ Augmentation}}} 
& \multirow{2}{*}{\makecell{Motor Imagery\\Classification}}
& BCI-IV-2A~\cite{tangermann2012review} & 250Hz & 22 & 4s & 9 & 5,184 & B-Acc & F1-W\\
& & SHU~\cite{ma2022large} & 250Hz & 32 & 4s & 25 & 11,988 & B-Acc & AUROC\\
\cmidrule{2-10}
& \makecell{Visual\\Decoding}
& Things-EEG~\cite{gifford2022large, grootswagers2022human} & 1,000Hz & 63 & 1s & 10 & 821,600 & 2-Way & Top-5\\
\midrule
\multirow{6}{*}{\textbf{\makecell{Clinical\\Monitoring}}} 
& \multirow{3}{*}{\makecell{Clinical Anomaly\\Detection}}
& TUEV~\cite{obeid2016temple} & 256Hz & 23 & 5s & 370 & 112,237 & B-Acc & F1-W\\
& & TUAB~\cite{obeid2016temple} & 250/256/512Hz & 23 & 10s & 2,383 & 409,083 & B-Acc & AUC-PR\\
& & Siena~\cite{pr8070846,goldberger2000physiobank} & 512Hz & 29 & 10s & 14 & 51,307 & B-Acc & AUC-PR\\
\cmidrule{2-10}
& \multirow{3}{*}{\makecell{Sleep Staging}}
& HMC~\cite{alvarez2021inter, goldberger2000physiobank} & 256Hz & 4 & 30s & 151 & 137,243 & B-Acc & F1-W\\
& & SHHS~\cite{zhang2018national, quan1997sleep} & 125Hz & 1 & 30s & 329 & 324,854 & B-Acc & F1-W\\
& & Sleep-EDF~\cite{kemp2000analysis, goldberger2000physiobank} & 100Hz & 2 & 30s & 78 & 414,961 & B-Acc & F1-W\\
\bottomrule
\end{tabular}
}
\end{table*}

\noindent \textbf{Clinical Anomaly Detection} focuses on identifying irregular patterns in brain activity to support the clinical diagnosis, monitoring of neurological conditions, and early identification of neurological disorders through EEG analysis.
We select \textbf{TUEV}~\cite{obeid2016temple}, \textbf{TUAB}~\cite{obeid2016temple}, and \textbf{Siena}~\cite{pr8070846,goldberger2000physiobank} datasets containing EEG signals collected from clinical patients from the Temple University Hospital (TUH) and Unit of Neurology and Neurophysiology of the University of Siena, respectively. 
TUEV includes 23-channel EEG recordings from 370 subjects and distinguishes EEG signals into 6 event types associated with seizures, including spike and/or sharp wave, generalized periodic epileptiform discharge, periodic lateralized epileptiform discharge, eye movement, artifact, and background. 
TUAB contains EEG recordings from 2383 subjects and classifies EEG recordings as clinically normal or abnormal activities. 
Siena includes 29-channel EEG recordings from 14 subjects, with recordings labeled as seizure or non-seizure events.

\noindent \textbf{Sleep Staging}, also referred to as sleep stage classification, identifies different sleep stages based on individual's EEG signals recorded during sleep, enabling early detection of sleep disorders and facilitating personalized healthcare solutions.
This task adopts commonly used sleep staging benchmarks, \textbf{HMC}~\cite{alvarez2021inter, goldberger2000physiobank}, \textbf{SHHS}~\cite{zhang2018national, quan1997sleep} and \textbf{Sleep-EDF}~\cite{kemp2000analysis, goldberger2000physiobank} datasets. These datasets categorize EEG signals into sleep stages in accordance with the AASM standard~\cite{aasm}: REM, N1, N2, N3, and Wake.
Specifically, HMC contains 4-channel EEG recordings from 151 subjects; SHHS provides 2-channel EEG recordings from 6,441 subjects; and Sleep-EDF includes 2-channel EEG recordings from 78 subjects.
Following~\cite{eldele2021attention}, we selected 329 healthy subjects from SHHS and used EEG signals recorded from the C4-A1 channel.


\subsection*{Evaluation Settings}
To provide a multifaceted analysis of brain foundation model's transferability to BCI tasks, we design three distinct evaluation settings: \textbf{cross-subject transfer}, \textbf{multi-subject adaptation}, and \textbf{few-shot transfer}. These settings systematically assess the model's generalization ability when adapting to downstream BCI tasks in diverse application scenarios. 


\noindent \textbf{Cross-Subject Transfer Setting.} This setting trains or fine-tunes models on EEG data from a group of subjects and tests them on new subjects whose data is explicitly excluded from training. This setting is to evaluate the generalization ability of models across a diverse range of individuals.
Given the inherent and pervasive inter-individual variability in brain anatomy, functional organization, and skull conductivity, effective cross-subject generalization is a critical challenge.

\noindent \textbf{Multi-Subject Adaptation Setting.} 
Models are trained on EEG data from multiple subjects and tested on data from the same subject cohort but from different recording sessions or trials. This setting assesses the model's robustness to intra-subject temporal variability in EEG signals, which is crucial for generalizing across sessions/trials for a group of seen subjects.
EEG signals are prone to time-related data distribution shifts due to unstable recording devices and individual mental state fluctuations~\cite{chen2022single, bigdely2008brain, huang2011framework, jobert2012guidelines}, leading to inconsistent decoding accuracy for the same subjects~\cite{saha2017evidence}.  
While previous work~\cite{Conformer, graph, chen2025adamgraph} has explored similar settings (\textit{e.g.}, within-subject, intra-subject), their data split strategies are inconsistent and limited to the single subject.
In this study, we propose a multi-subject adaptation setting for emotion recognition and motor imagery tasks for fair comparison.


%

\noindent \textbf{Few-Shot Transfer Setting.} This setting is to evaluate the capacity of foundation models to swiftly adapt to new tasks with limited data and minimal labels. Following the practice of previous multi-modal foundation models~\cite{CLIP, COOP}, we sample a small number of samples from each category for each subject to create the support set, used for training or fine-tuning. Then the model is evaluated on the query set from the remaining samples.
Specifically, the support set consists of a proportion of randomly sampled instances for each category, with the proportion $k$ ranging from $[0.02, 0.05, 0.3, 0.5]$. The results are averaged from three runs. This setting simulates the scenario where a new task/user has only a small amount of calibration data, which is vital for real-world BCI deployment where gathering extensive data is often impractical and expensive. 

\subsection*{Task Adaptation Pipeline}

\subsubsection*{Standardized Data Preprocessing}

As EEG signals are inherently noisy and complex, data preprocessing steps are critical to enhance signal quality and mitigate artifacts. Here, we develop a standardized data preprocessing pipeline including band-pass filtering, notch filtering, resampling, and normalization, that can be universally applied in various BCI tasks and decoding models.

\noindent\textbf{Band-Pass Filtering.}
EEG signals span 0 Hz to hundreds of Hz, but among which only specific frequency bands are pertinent to the task. Following previous practice~\cite{labram}, we employ a 0.1–75 Hz band-pass filter, preserving crucial task-relevant frequency components like $\delta$ (0.5–4 Hz), $\theta$ (4–8 Hz), $\alpha$ (8–12 Hz), $\beta$ (12–30 Hz), and $\gamma$ (30–50 Hz),  while suppressing low-frequency drifts and high-frequency noise.

\noindent\textbf{Notch Filtering.}
To eliminate power line interference, notch filtering was applied to the EEG signals in this study. A Fast Fourier Transform (FFT) is conducted on the raw EEG data to identify the power-line frequency at the collection site from the frequency spectrum.

\noindent\textbf{Resampling.}
Raw EEG signals are typically recorded at high sampling rates (\textit{e.g.}, 1000 Hz or higher), leading to data redundancy and increased computational overhead. Additionally, the sampling rates of different datasets often vary. Therefore, we resampled the EEG signals to align with the sampling rate configured in the evaluated model.

\noindent\textbf{Normalization.}
Data normalization is essential to accommodate varying data distributions across EEG datasets. We applied \textbf{z-score normalization} as the standard technique for most datasets, which normalizes each sample using global statistics across all trials to conform to a normal distribution.
For SHU dataset, which is prone to outliers, we used \textbf{95-percentile normalization} to enhance robustness, which normalizes data points by dividing them by the 95-th percentile of absolute amplitude per channel in each trial, thus mitigating the impact of outliers.
For Things-EEG dataset, we directly rescaled the signal to ensure that most signal values fall within the range of approximately $[-1,1]$ following previous practice~\cite{labram}.




\subsubsection*{Adaptation Strategy}

AdaBrain-Bench does not constrain the adaptation strategy. Here we use two commonly-used adaptation strategies for fine-tuning: \textbf{full fine-tuning} and \textbf{linear probing}. 
The full fine-tuning approach updates all model parameters when adapting to new tasks. Although computationally intensive, this method generally yields superior performance with sufficient training data~\cite{kornblith2019better} when the upstream and downstream datasets differ significantly, as it facilitates task-specific feature learning across all layers of the network. Thus we adopt full fine-tuning as the primary transfer strategy.
It is also common to freeze the pretrained foundation models and fine-tune additional linear layers to adapt to downstream tasks in transfer learning, a method referred to as linear probe strategy. While computationally efficient, linear probe demands more transferable representations from EEG foundation models. 

\subsubsection*{Task Heads}
We developed several specialized task heads to handle diverse task paradigms, including classification, regression, and retrieval. These task heads are lightweight and can be seamlessly attached to models in a plug-and-play manner.


For classification tasks, the output embeddings of the model first undergo dimension reduction using its original methodologies (average pooling or linear reduction). The resulting embeddings are then flattened into one-dimensional vectors before being fed into a linear classification layer for final prediction.
%
For regression-based decoding, the model's output embeddings undergo average pooling before being processed by a three-layer linear regression network. Each linear layer is sequentially followed by an ELU activation and a dropout layer.
%
For the retrieval task (i.e., visual decoding on Things-EEG), the model outputs are first uniformly flattened and then are processed through a weight-normalized linear projection layer.
To evaluate visual semantic decoding performance, we adopt the contrastive learning framework from the previous work~\cite{ATM} to align projected EEG embeddings with corresponding image features. The EEG-to-image retrieval is subsequently conducted.

\noindent \textbf{Other Experimental Details.}
We use binary cross-entropy (BCE) loss for binary classification, cross-entropy (CE) loss for multiclass classification, mean squared error (MSE) for regression, and contrastive loss for retrieval. We perform a learning rate grid search over $[1e-3, 5e-4, 1e-4, 5e-4, 1e-5]$ for each adaptation strategy/setting combination to determine the optimal value. The batch size is set as 64. We adopted the AdamW optimizer, with the weight decay set as 0.05. The number of training epoch is set as 50. The experiments are conducted using Python 3.10.13 and PyTorch 2.4.0 with CUDA 11.8 support. 
Notably, for foundation models (\textit{e.g.,} BIOT and EEGPT) unable to directly decode downstream datasets due to channel system incompatibilities, we applied a $1\times 1$ channel-wise convolutional layer to ensure compatibility with input requirements.



\subsubsection*{Evaluation Metrics}
Our benchmark employs various evaluation metrics to assess the decoding performance for different kinds of tasks: binary classification, multiclass classification, regression and retrieval tasks. Table~\ref{tab:bci-datasets} summarizes the evaluation metrics corresponding to each dataset in AdaBrain-Bench.

\noindent \textbf{Accuracy Metrics.}
For \textit{binary classification} tasks, different evaluation metrics are adopted based on the characteristics of each dataset. For balanced binary classification tasks (\textit{e.g.}, EEGMAT and SHU), we report the Balanced Accuracy (\textbf{B-Acc}) and the Area Under the Receiver Operating Characteristic Curve (\textbf{AUROC}). B-Acc is defined as the average recall across all classes, while AUROC quantifies the model’s overall discriminative ability across different classification thresholds. For imbalanced classification tasks (\textit{e.g.}, TUAB and Siena), where the positive class is of particular importance, we report \textbf{B-Acc} and the Area Under the Precision-Recall Curve (\textbf{AUC-PR}), as AUC-PR more effectively reflects the model’s performance on the minority (positive) class.
For \textit{multiclass classification} tasks (\textit{i.e.}, SEED, SEED-IV, EEGMAT, BCI-IV-2A, TUEV, HMC, SHHS, and Sleep-EDF), we use B-Acc and the Weighted F1-Score (\textbf{F1-W}) as the metric. F1-W evaluates model performance by calculating a weighted average of each class's F1 score, where weights are determined by class sample proportion to alleviate class imbalance issue. 
For \textit{regression tasks} (\textit{i.e.}, SEED-VIG), we adopt Pearson Correlation Coefficient (\textbf{Corr}) and \textbf{R2 score} as the metric. Corr measures the linear relationship between predicted and true values, while R2 represents the proportion of variance in the dependent variable explained by the decoding model.
For \textit{retrieval tasks} (\textit{i.e.}, Things-EEG), we use \textbf{2-Way} and \textbf{Top-5} accuracy as the metric, following the previous practice~\cite{ATM}. The N-way accuracy is computed by constructing a set containing the target sample and $N-1$ distractor samples from the test set, assessing the model's ability to identify the correct target among $N$ categories. Top-5 accuracy measures the frequency that the true target image is among the top 5 retrieved results, based on the corresponding EEG signals.

\noindent \textbf{Transferability Metric.}
To further quantify the transferability of foundation models, we introduce the \textit{\textbf{transfer score} (TS)}, which is defined as the combination of the relative gain of a pretrained foundation model over its trained-from-scratch counterpart and the relative improvement to the maximum possible gain:
\begin{equation}
\text{TS} = \lambda\cdot\frac{P_{\text{prt}}-P_{\text{scr}}}{P_{\text{scr}}} + (1-\lambda)\cdot\frac{P_{\text{prt}}- P_{\text{scr}}}{P_{\text{oracle}} - P_{\text{scr}}},
\end{equation}
where $P_{\text{prt}}$ denotes the performance of the adapted foundation model, $P_{\text{scr}}$ denotes the performance achieved by training the same model from scratch. $P_{\text{oracle}}$ denotes the approximate upper bound, which is set as 1 for the cross-subject transfer setting and the full-data performance for the few-shot transfer setting. $\lambda$ is set as default 0.5 in the experiment.

\begin{table*}[tbp]
\centering
\caption{\textbf{Summary of recently published brain foundation models.}}
\label{tab:summary-model}
\renewcommand{\arraystretch}{1.0}
\setlength{\tabcolsep}{1pt}
\resizebox{0.99\textwidth}{!}{
\begin{tabular}{c|ccccccc}
\toprule
\textbf{Model} & \textbf{\#Params.} & \textbf{Architecture} & \makecell{\textbf{Pretraining} \\ \textbf{Strategy}} & \makecell{\textbf{Pretraining Data Size} \\ \textbf{(\#channels * h)}} & \makecell{\textbf{Pretraining} \\ \textbf{Signal Duration}}  & \textbf{Pre-training Datasets} & \makecell{\textbf{\#Supported}\\ \textbf{Channels}} \\
\midrule
\textbf{BIOT}~\cite{yang2023biot} & 3.2M & Linear Transformer & 
\makecell{Self-supervised\\Contrastive Learning} & 354,625 & 59,304h & 
\makecell{PREST, SHHS, CHB-MIT, \\ IIIC Seizure, TUAB \& TUEV} & 18 \\
\midrule
\textbf{EEGPT}~\cite{eegpt} & 25M & CNN \& Transformer & 
\makecell{Masked Signal\\Reconstruction} & 11,476 & 198h & 
\makecell{PhysioNet-MI, HGD, TSU, \\ SEED \& M3CV} & 58 \\
\midrule
\textbf{LaBraM}~\cite{labram} & 5.8M & CNN \& Transformer & 
\makecell{Masked Signal\\Reconstruction} & 76,801 &  2,535h & 
\makecell{BCI-IV-1, Emobrain, SPIS, \\ Grasp and Lift EEG, Inria P300, \\ SEED series, Siena, \\ TUEP, TUSZ \& others} & 136 \\
\midrule
\textbf{CBraMod}~\cite{wang2024cbramod} & 4.0M & CNN \& Transformer & 
\makecell{Masked Signal\\Reconstruction} & 175,678 & 9,246h & \makecell{TUAB, TUAR \\ TUEP, TUEV \\ TUSE \& TUSL} & Flexible \\
\bottomrule
\end{tabular}
}
\end{table*}

%




\subsubsection*{Models}

We considered several open-sourced pre-trained EEG foundation models to be evaluated in our benchmark, including LaBraM~\cite{labram}, EEGPT~\cite{eegpt}, CBraMod~\cite{wang2024cbramod}, and BIOT~\cite{yang2023biot}. Next we provide details of these methods and summary in Table~\ref{tab:summary-model}. Apart from these EEG foundation models, we also included traditional supervised models with state-of-the-art performance as baselines, including EEGNet~\cite{Lawhern}, LDMA~\cite{miao2023lmda}, ST-Tran~\cite{Transformer}, and Conformer~\cite{Conformer}.

\noindent \textbf{BIOT~\cite{yang2023biot}.}
BIOT adopts linear transformer architecture that facilitates effective cross-data learning by uniform EEG tokenization. The tokenization involves dividing EEG signals into fixed-length segments. The segment embeddings along with spatial and temporal position embeddings are re-arranged into token sequences. BIOT conducts self-supervised contrastive learning between masked and unmasked signal embeddings. We utilize the BIOT version pretrained on six clinical datasets related to seizure and sleep, supporting 18 pre-defined channels.


\noindent \textbf{EEGPT~\cite{eegpt}.}
EEGPT is a transformer-based foundation model adopting mask-based self-supervised learning. The pretraining involves a mask-based reconstruction strategy and a spatio-temporal representation alignment strategy to learn generic neural representations from unlabeled EEG data. EEGPT was pretrained on 246 hours of EEG signals from five datasets from various tasks, rendering a large-scale foundation model with 25 million parameters and 58 supported channels.




\noindent \textbf{LaBraM~\cite{labram}.}
LaBraM is a pretrained transformer model that employs a codebook-based neural tokenizer. The tokenizer is trained by vector-quantized neural spectrum prediction to convert raw EEG signals into discrete neural code. The code is then used for masked EEG pretraining, allowing the model to capture generalizable neural representations. LaBraM was pretrained on about 2,500 hours of various types of EEG signals from around 20 datasets, supporting 137 EEG channels as input.

\noindent \textbf{CBraMod~\cite{wang2024cbramod}.}
CBraMod refines LaBraM by implementing a criss-cross transformer that separately models spatial and temporal dependencies separately via parallel attention mechanisms. It also utilizes dynamic positional encoding conditioned on EEG signals, enabling adaptation to arbitrary EEG channels. The model was pretrained on approximately 27,000 hours of EEG signals from the Temple University Hospital EEG Corpus (TUEG)~\cite{obeid2016temple}.

\section*{Data Availability}

All the datasets used in this benchmark are publicly available at the  \href{https://github.com/Jamine-W/AdaBrain-Bench}{GitHub repository} along with instructions for data splitting and preprocessing to reproduce the reported results.

\section*{Code Availability}

The code associated with this work is available in this \href{https://github.com/Jamine-W/AdaBrain-Bench}{GitHub repository}  with a MIT License.

\bibliography{neurobench}





\end{document}